\documentclass{article}
\usepackage{spconf,amsmath,graphicx}
\usepackage[hyphens]{url}
\usepackage{amsmath}
\usepackage{paralist}
\usepackage{booktabs}
\usepackage[dvipsnames]{xcolor}
\usepackage{microtype}
\usepackage{ragged2e}
\PassOptionsToPackage{hyphens}{url}
\usepackage{caption}
\usepackage{subcaption}
\RequirePackage{xcolor}
\usepackage{multirow}
\usepackage[hyphens]{url}
\usepackage[pagebackref=false,breaklinks=true,colorlinks,bookmarks=false]{hyperref}

\def\ie{{\em i.e}}
\def\eg{{\em e.g}}

\def\eg{\emph{e.g}}

\def\ie{\emph{i.e}}

\def\longversion{1}

\begin{document}

\title{Exposing GAN-generated Faces Using \\ Inconsistent Corneal Specular Highlights}

\name{Shu Hu, Yuezun Li, and Siwei Lyu}
\address{Computer Science and Engineering \\University at Buffalo, State University of New York, USA\\{\tt\{shuhu,yuezunli,siweilyu\}@buffalo.edu}}

\maketitle

\begin{abstract}
Sophisticated generative adversary network (GAN) models are now able to synthesize highly realistic human faces that are difficult to discern from real ones visually. In this work, we show that GAN synthesized faces can be exposed with the inconsistent corneal specular highlights between two eyes. The inconsistency is caused by the lack of physical/physiological constraints in the GAN models. We show that such artifacts exist widely in high-quality GAN synthesized faces and further describe an automatic method to extract and compare corneal specular highlights from two eyes. Qualitative and quantitative evaluations of our method suggest its simplicity and effectiveness in distinguishing GAN synthesized faces.
\end{abstract}

\section{Introduction}

\begin{figure}[t]
    \centering
    \includegraphics[width=0.5\textwidth]{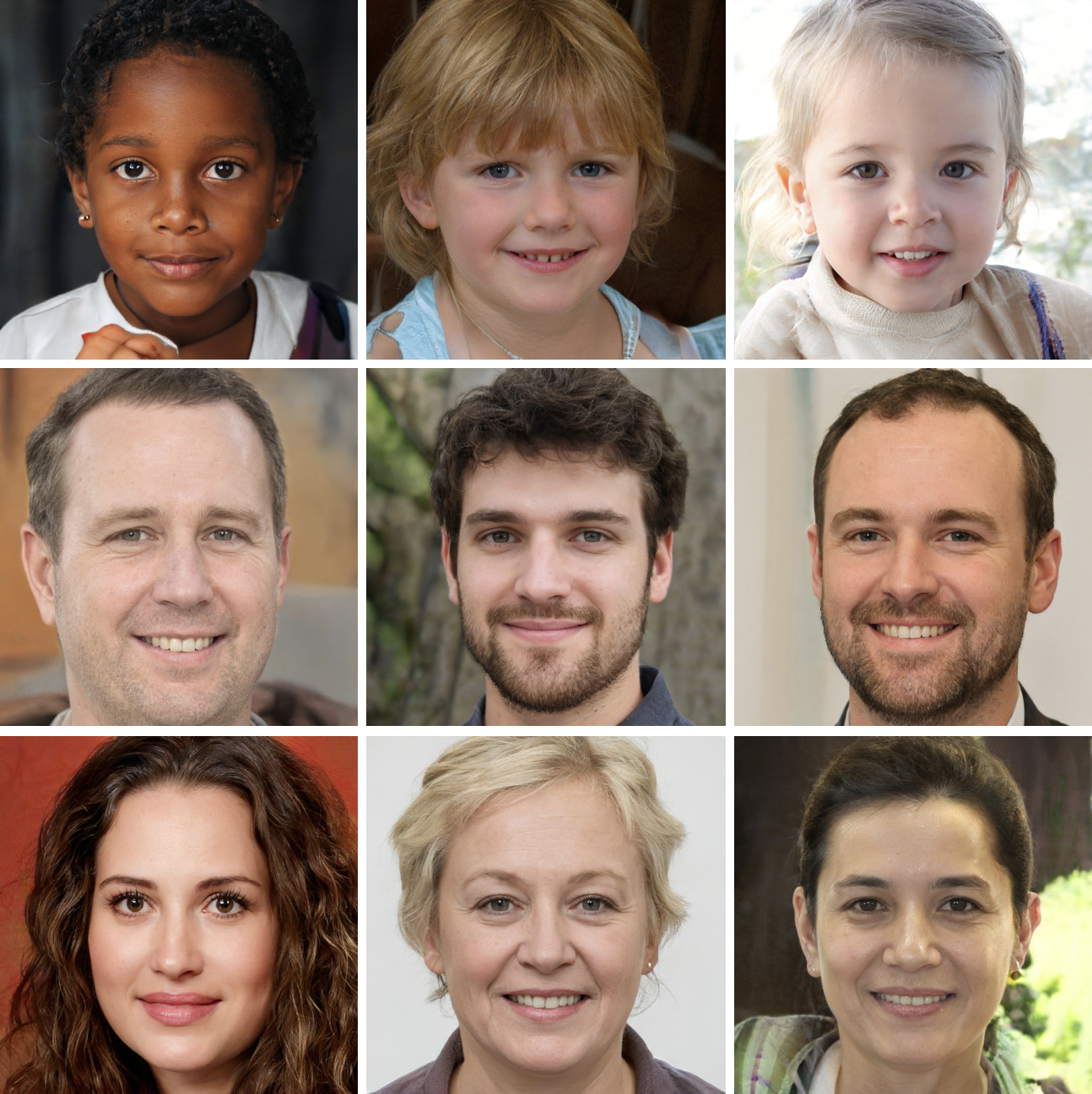}
    \vspace{-2em}
    \caption{\em Examples of GAN synthesized images of realistic human faces. These images are obtained from \url{http://thispersondoesnotexist.com} generated with the StyleGAN2 model \cite{karras2020analyzing}.}
    \vspace{-1.5em}
    \label{fig:gan-example}
\end{figure}

The rapid advancements of the AI technology, the easier access to large volume of online personal media, and the increasing availability of high-throughput computing hardware have revolutionized the manipulation and synthesis of digital audios, images, and videos. A quintessential example of the AI synthesized media are the highly realistic human faces generated using the generative adversary network (GAN) models \cite{goodfellow2014generative,karras2017progressive,karras2019style,karras2020analyzing}, Figure \ref{fig:gan-example}. As the GAN-synthesized faces have passed the ``uncanny valley'' and are challenging to distinguish from images of real human faces, they quickly become a new form of online disinformation. In particular, GAN-synthesized faces have been used as profile images for fake social media accounts to lure or deceive unaware users \cite{theverge,cnn1,cnn2,reuters}. 

\begin{figure}[t]
    \centering
    \includegraphics[width=.45\textwidth]{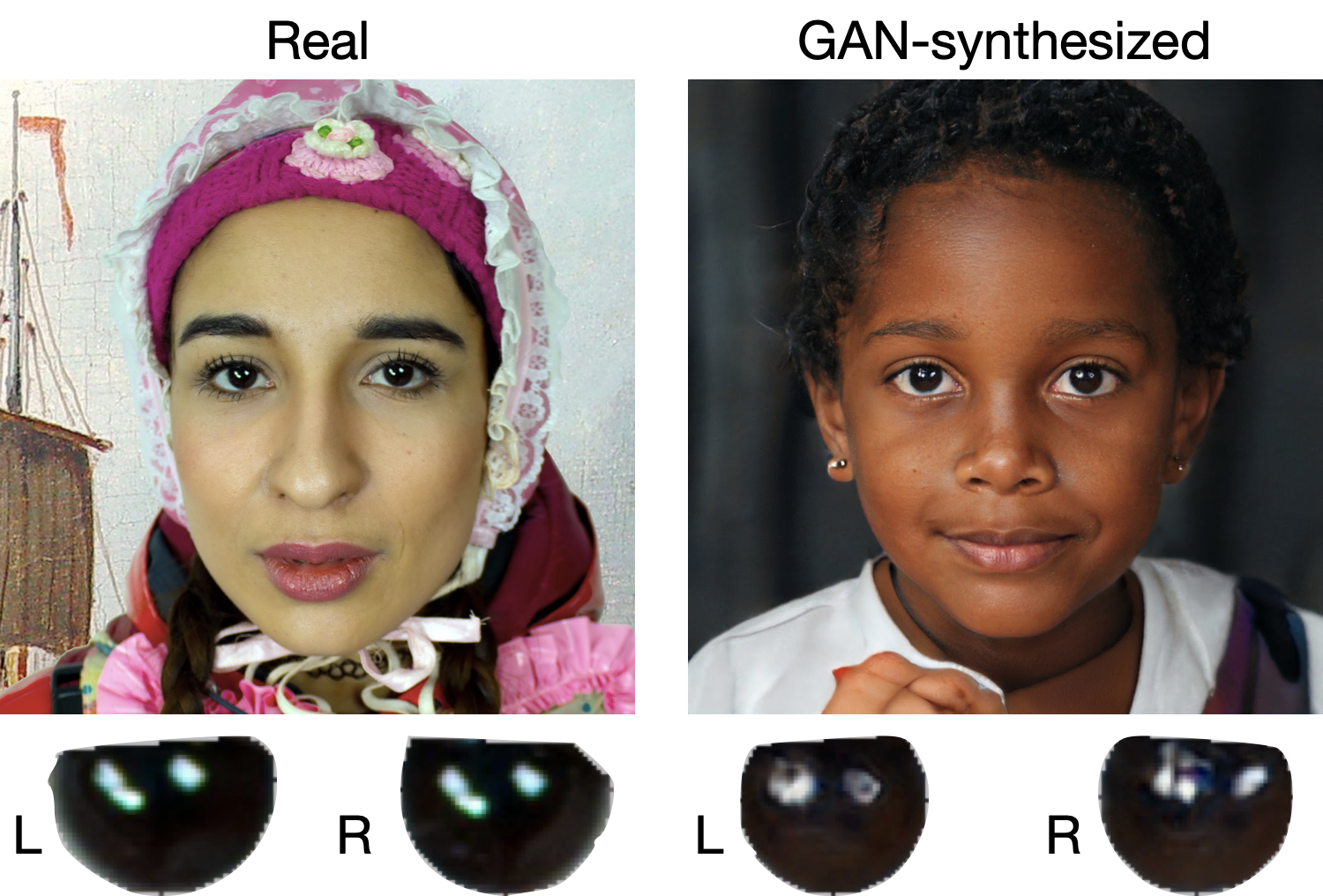}
    \vspace{-1em}
    \caption{\em Corneal specular highlights for a real human face (left) and a GAN-synthesized face (right). The corneal regions are isolated and scaled for better visibility. Note that the corneal specular highlights for the real face have strong similarities while those for the GAN-synthesized face are different.}
    \vspace{-1em}
    \label{fig:compare}
\end{figure}
Correspondingly, there is a rapid development of detection methods targeting at GAN synthesized faces \cite{yang2018exposing,marra2019gans}. The majority of GAN-synthesized image detection methods are based on extracting signal level cues then train classifiers such as SVMs or deep neural networks to distinguish them from real images. Although high performance has been reported using these methods, they also suffer from some common drawbacks, including the lack of interpretability of the detection results, low robustness to laundering operations and adversarial attacks \cite{calini_farid}, and poor generalization across different synthesis methods. A different type of detection methods take advantage of the inadequacy of the GAN synthesis models in representing the more semantic aspects of the human faces and their interactions with the physical world \cite{yang2019exposinggan,yang2018exposing,li2018detection,matern2019exploiting}. Such physiological/physical based detection methods are more robust to adversarial attacks, and afford intuitive interpretations. 

In this work, we propose a new physiological/physical based detection method of GAN-synthesized faces that uses the inconsistency of the corneal specular highlights between the two synthesized eyes. The corneal specular highlights are the images of light emitting or reflecting objects in the environment at the time of capture on the surface of the cornea. When the subject's eyes look straight at the camera and the light sources or reflections in the surrounding environment are relatively far away from the subject (\ie, the ``portrait setting''), the two eyes see the same scene and their corresponding corneal specular highlights exhibit strong similarities (Figure \ref{fig:compare}, left image). We observe that GAN-synthesized faces also comply with the portrait setting (Figure \ref{fig:gan-example}), possibly inherited from the real face images that are used to train the GAN models. However, we also note the striking inconsistencies between the corneal specular highlights of the two eyes (Figure \ref{fig:compare}, right image). Our method automatically extracts and aligns the corneal specular highlights from two eyes and compare their similarity. Our experiments show that there is a clear separation between the distribution of the similarity scores of the real and GAN synthesized faces, which can be used as a quantitative feature to differentiate them.  

\section{Background}

\begin{figure}[t]
    \centering
    \includegraphics[width=.45\textwidth]{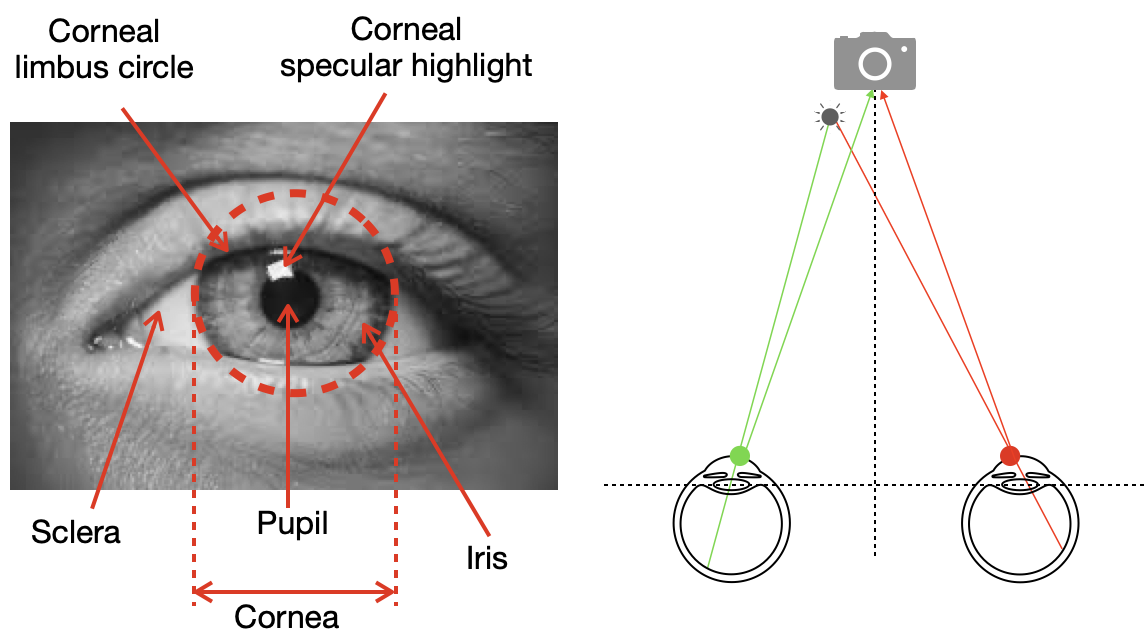}
    \vspace{-1em}
    \caption{\em (left) Anatomy of a human eye. (right) The portrait setting with the corneal specular highlights.}
    \vspace{-1em}
    \label{fig:eye}
\end{figure}

\noindent {\bf Anatomy of Human Eyes}. The human eye provides the optics and photo-reception for the visual system. Figure \ref{fig:eye} shows the main anatomic parts of a human eye. The center of an eye are iris and the pupil. The transparent cornea is the outer layer that covers the iris and dissolves into the white sclera at the circular band known as the corneal limbus. The cornea has a spherical shape and its surface exhibits mirror-like reflection characteristics, which generates the corneal specular highlights when illuminated by light emitted or reflected in the environment at the time of capture.

\begin{figure*}[t!]
    \centering
    \includegraphics[width=.9\textwidth]{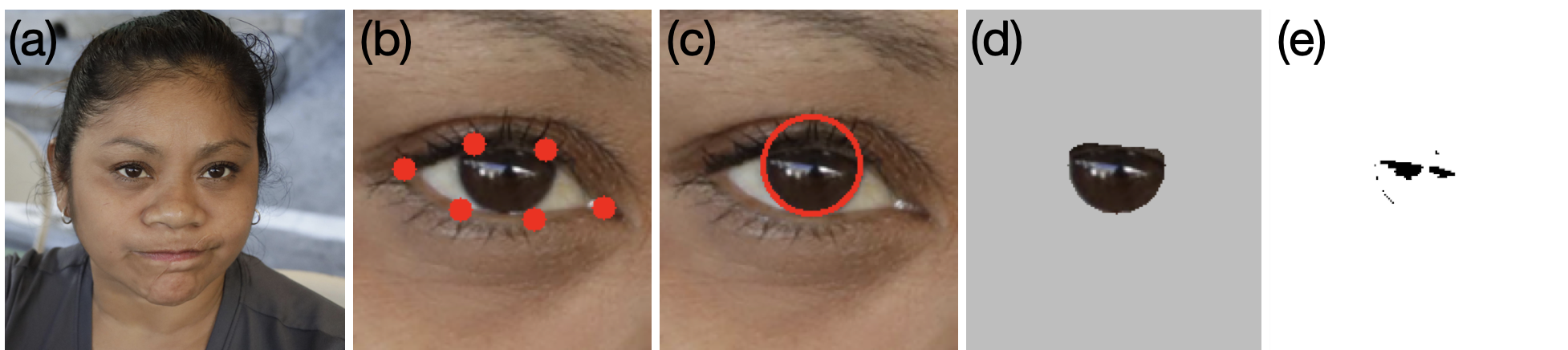}
    \vspace{-1em}
    \caption{\em Overall process to obtain corneal specular highlight. (a) The input high-resolution face image. (b) Detection of facial landmarks around the eyes. (c) Hough circle detection of the corneal area. (d) Intersection of the eye region and circular corneal region. (e) Extracted corneal specular highlight area.}
    \vspace{-1em}
    \label{fig:pipeline}
\end{figure*}  


\medskip
\noindent {\bf GAN Synthesis of Human Faces}. 
A series of recent works known as StyleGANs \cite{karras2017progressive, karras2019style, karras2020analyzing} have demonstrated the superior capacity of GAN models \cite{goodfellow2014generative} trained on large sets of real human faces in generating high-resolution realistic human faces. A GAN model consists of two neural networks trained in tandem. The generator takes random noises as input and synthesizes an image, and the discriminator aims to differentiate synthesized images from the real ones. In training the two networks compete with each other: the generator aims to create more realistic images to defeat the discriminator, while the discriminator aims to improve the accuracy in differentiating the two types of images. The training ends when the two networks reach an equilibrium.

Albeit the successes, GAN-synthesized faces are not perfect. Early StyleGAN model was shown to generate faces with asymmetric faces \cite{yang2019exposinggan} and inconsistent eye colors \cite{matern2019exploiting}. However, the more recent StyleGAN2 model \cite{karras2020analyzing} further improves the synthesis quality and eliminate such artifacts. However, visible artifacts and inconsistencies can still be observed in the  background, the hair, and the eye regions. One fundamental reason for the existence of such global and semantic artifacts in GAN synthesized faces is due to their lack of understanding of human face anatomy, especially the geometrical relations among the facial parts. 



\section{related works}

Methods detecting GAN-synthesized faces fall into three categories. Those in the first category focus on signal traces or artifacts left by the GAN synthesis model. For example, earlier works, \eg, \cite{mccloskey2018detecting,li2018detection}, use color differences of first generation of GAN images. As color difference can be easily fixed, more sophisticated detection methods, \eg, \cite{marra2019gans,yu2019attributing}, seek more abstract signal-level traces or fingerprints in the noise residuals to differentiate GAN-synthesized faces. More recent works such as \cite{zhang2019detecting,frank2020leveraging,durall2020watch} extend the analysis to the frequency domain, where the upsampling step in the GAN generation leaves specific artifacts. The second category of GAN synthesized face detection methods are of data-driven nature \cite{marra2019incremental,goebel2020detection,wang2020cnn,liu2020global,hulzebosch2020detecting}, where a deep neural network model is trained and employed to classify real and GAN-synthesized faces. Methods of the third category look for physical/physiological inconsistencies by GAN models. The work in \cite{yang2019exposinggan} distinguish GAN-synthesized faces by analyzing the distributions of facial landmarks, and \cite{yang2018exposing} {exposes the fake videos by detecting inconsistent head poses}. The method in \cite{matern2019exploiting} further inspect more visual aspects to expose GAN synthesized faces. Such physiological/physical based detection methods are more robust to adversarial attacks, and afford intuitive interpretations. 


Because of the unique geometrical regularity, the corneal region of the eyes have been used in the forensic analysis of digital images. The work of \cite{johnson06specular} estimates the internal camera parameters and light source directions from the perspective distortion of the corneal limbus and the locations of the corneal specular highlights of two eyes, which are used to reveal digital images composed from real human faces photographed under different illumination. 
The work of \cite{matern2019exploiting} identifies early generations of GAN synthesized faces \cite{karras2017progressive} by noticing that they may have inconsistent iris colors, and the specular reflection from the eyes is either missing or appear simplified as a white blob. However, such inconsistencies have been largely improved in the current state-of-the-art GAN synthesis models (\eg, \cite{karras2020analyzing}), see examples in Figure \ref{fig:gan-example}.

\begin{figure*}[t!]
    \centering
    Images of real human eyes \\
    \includegraphics[width=.9\textwidth]{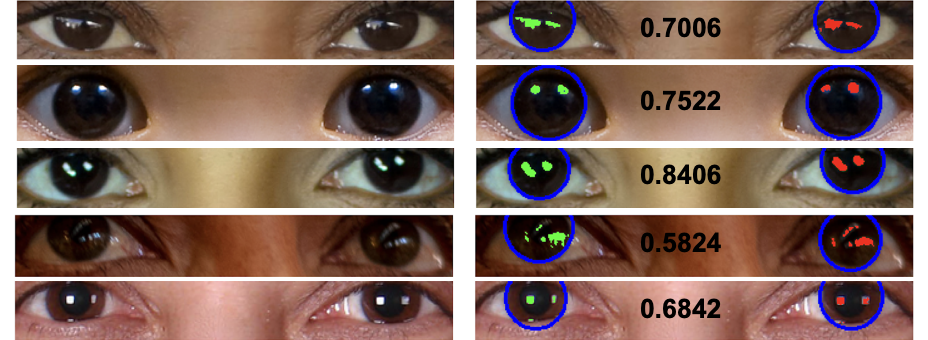} \\
    Images of GAN synthesized human eyes \\
    \includegraphics[width=.9\textwidth]{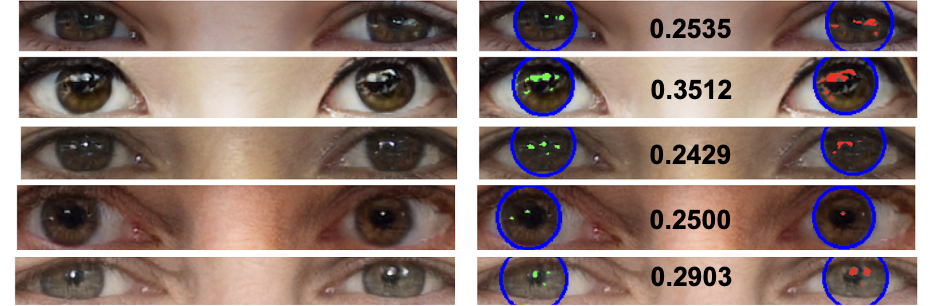}
    \vspace{-1em}
    \caption{\em Corneal specular highlights from real human eyes (top) and GAN generated human faces (bottom). The right column corresponds to the detected corneal region (blue) and the specular highlights of two eyes (green and red). The IoU scores of the two corneal specular highlights are shown alongside the detections.}
    \label{fig:eye_examples}
    \vspace{-1.em}
\end{figure*}

\section{Method}

In this work, we explore the use of corneal specular highlight as a cue to expose GAN synthesized human faces. The rationale of our method can be understood as follows.  In an image of a real human face captured by a camera, the corneal specular highlights of the two eyes are related as they are the results of the same light environment. Specifically, they are related by a transform that is determined by (1) the anatomic parameters of the two eyes including the distance between the centers of the pupils and the diameters of the corneal limbus; (2) the poses of the two eyeballs relative to the camera coordinate system, \ie, their relative location as a result of head orientation; and (3) the location and distance of the light sources to the two eyes, measured in the camera coordinates. 

Under the following conditions, which we term as the {\em portrait setting} as it is often the case in practice when shooting closeup portrait photographs, the corneal specular highlights of the two eyes have approximately the same shape. To be more specific, what we mean by a portrait setting consists of the following conditions, which is also graphically illustrated in the right panel of Figure \ref{fig:eye}.
\begin{compactitem} \itemsep 0em
\item The two eyes have a frontal pose, \ie, the line connecting the center of the eyeballs is parallel to the camera.
\item The eyes are distant from the light or reflection source.
\item All light sources or reflectors in the environment are visible to both eyes.
\end{compactitem}

To highlight such artifacts and quantify them as a cue to expose GAN synthesized faces, we develop a method to automatically compare the corneal specular highlights of the two eyes and evaluate their similarity. Figure \ref{fig:pipeline} illustrates major steps of our analysis for an input image. We first run a face detector to locate the face, followed by a landmark extractor to obtain landmarks (Figure \ref{fig:pipeline}(b)), which are important locations such as the face contour, tips of the eyes, mouth, nose, and eyebrows, on faces that carry important shape information. The regions corresponding to the two eyes are properly cropped out using the landmarks. We then extract the corneal limbus, which affords a circular form under the portrait setting. To this end, we first apply a Canny edge detector followed by the Hough transform to find the corneal limbus (Figure \ref{fig:pipeline}(c)), and  use its intersection with the eye region provided by the landmarks as the corneal region (Figure \ref{fig:pipeline}(d)).


\begin{figure}[htbp]
\begin{subfigure}[t]{0.49\linewidth}
    \includegraphics[width=\linewidth]{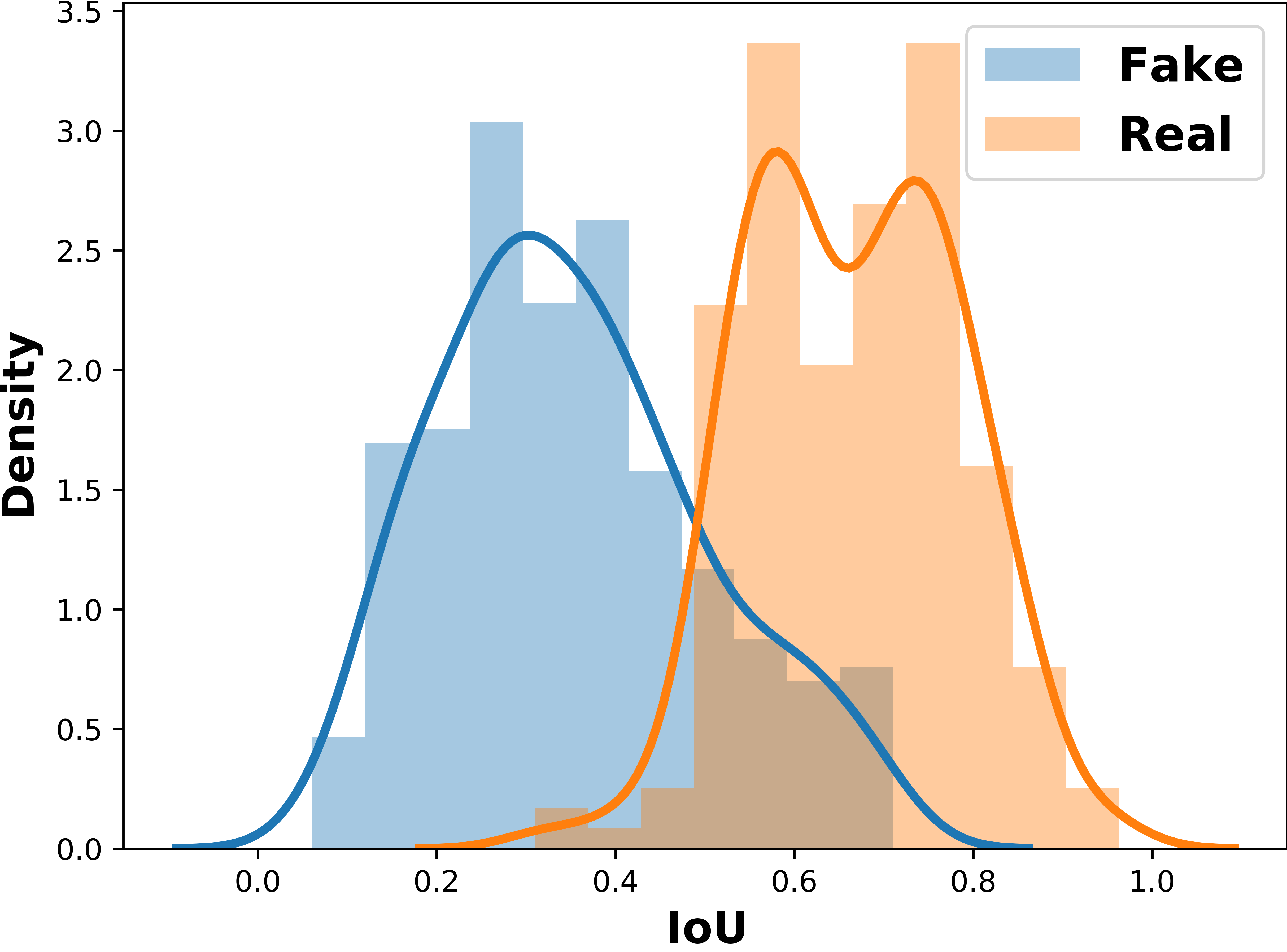}
    \caption{}
\end{subfigure}%
    \hfill%
\begin{subfigure}[t]{0.49\linewidth}
    \includegraphics[width=\linewidth]{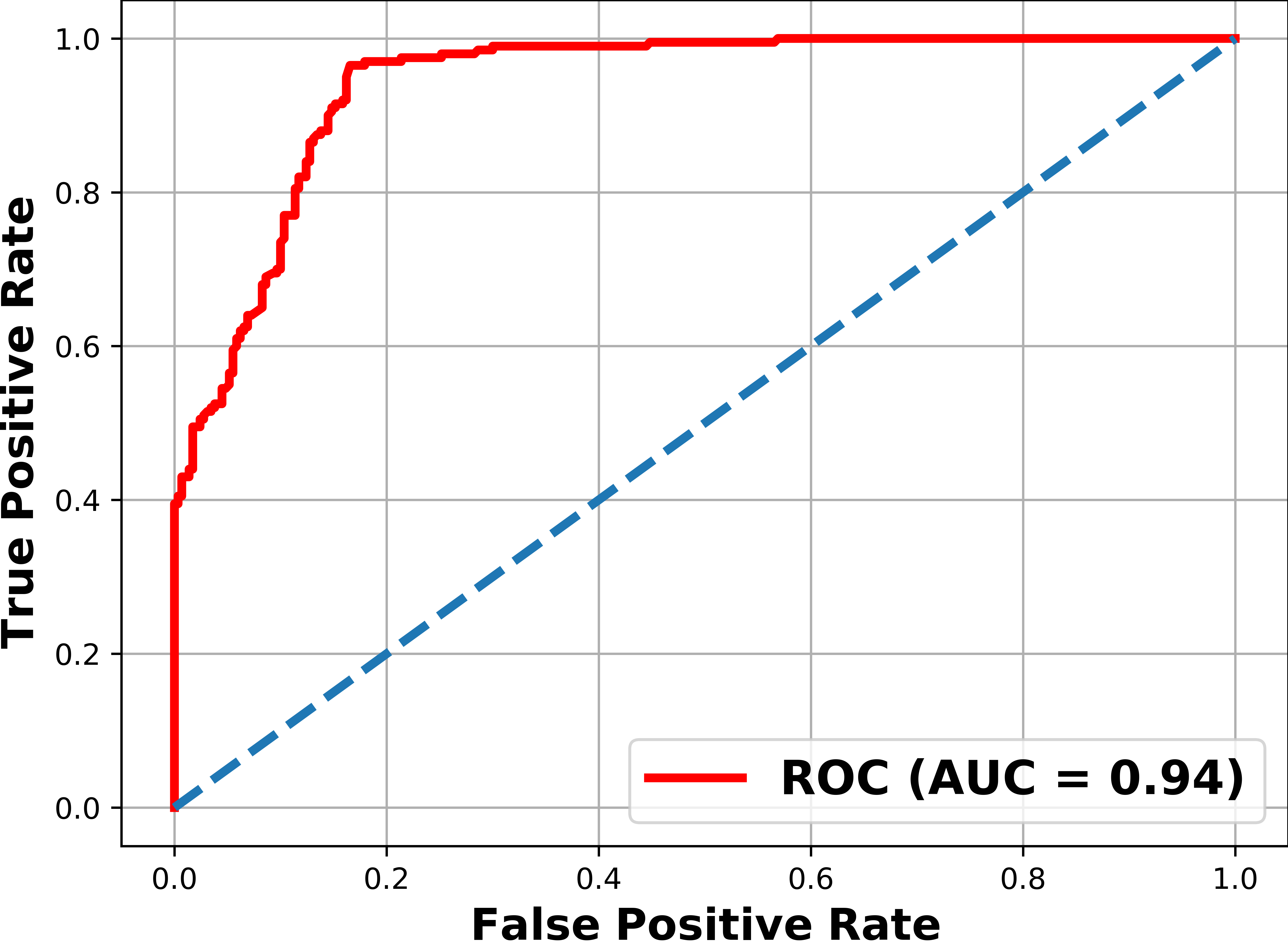}
    \caption{}
\end{subfigure}
\vspace{-1em}
\caption{\em (a) Distributions of the IoU scores between the detected corneal specular highlights of two eyes for real and GAN synthesized faces. (b) The ROC curve based on the IoU scores.}
\vspace{-2.5em}
\label{fig:histo}
\end{figure}

We then separate the corneal specular highlights using an adaptive image thresholding method \cite{yen1995new}. Because the specular highlights tend to have brighter intensities than the background iris, we keep only pixel locations above the adaptive threshold (Figure \ref{fig:pipeline}(e)). We align the extracted corneal specular highlights of the two eyes (denoted as $R_L$ and $R_R$) with a translation, and use their IoU scores, ${\frac{|R_L \cap R_R|}{|R_L \cup R_R|}}$, as a similarity metric. The IoU score takes range in $[0,1]$ with smaller value suggesting lower similarity of $R_L$ and $R_R$, and hence more likely the face is created with a GAN model.

\section{Experiments}

The images of real human eyes are obtained from the Flickr-Faces-HQ (FFHQ) dataset \cite{karras2019style}, and the GAN synthesized human faces are from \url{http://thispersondoesnotexist.com}, which are created by the StyleGAN2 method \cite{karras2020analyzing}. The images have resolution of $1,024 \times 1,024$ pixels. We use the face detector and landmark extractor provided in DLib \cite{dlib09}, and the Canny edge detector and Hough transform are from scikit-image \cite{van2014scikit}.

Figure \ref{fig:eye_examples} shows examples of the analysis results for images of both real and GAN-synthesized human eyes. As described in the previous section, real human eyes captured by a camera under the portrait setting exhibit strong resemblance between the corneal specular highlights of the two eyes, which are reflected by the higher IoU scores. On the other hand, the corneal specular highlights of the two GAN synthesized eyes may exhibit various types of inconsistencies, such as different numbers, different geometric shapes, or different relative locations of specular highlight regions of the two eyes. These artifacts lead to significantly lower IoU scores. Figure \ref{fig:histo} (a) shows the distributions of the IoU scores of two eyes' corneal specular highlights for the real images and GAN generated images we collected. Consistent with the visual examples, there is a clear separation between the distributions, indicating that consistency of corneal specular highlights is an effective measure differentiating real and GAN generated faces. We also show the {\em receiver operating characteristic} (ROC) curve in Figure \ref{fig:histo} (b), which corresponds to an AUC (Area under the ROC curve) score of 0.94, indicating that corneal specular highlights are effective to identify GAN synthesized faces. 

\section{Discussion}

In this work, we show that GAN synthesized faces can be exposed with the inconsistent corneal specular highlights between two eyes. Although inconsistencies of specular patterns can be fixed with manual post-processing, it is expected to be non-trivial. Our method has several limitations. We only compare pixel difference without consider inconsistencies in geometry and scene. Also, when the portrait setting is not obeyed, we may have false positives, \eg, when light source is very close to the subject or a peripheral light source that is not visible in both eyes. It does not apply to images where specular patterns are not present. In the future, we will investigate these aspects and further improve the effectiveness of our method.

\newpage
{\small
\bibliographystyle{IEEEbib}
{\raggedright
\bibliography{refs}
}
}

\if\longversion1
\begin{figure}[t]
     \centering
     \includegraphics[width=.5\textwidth]{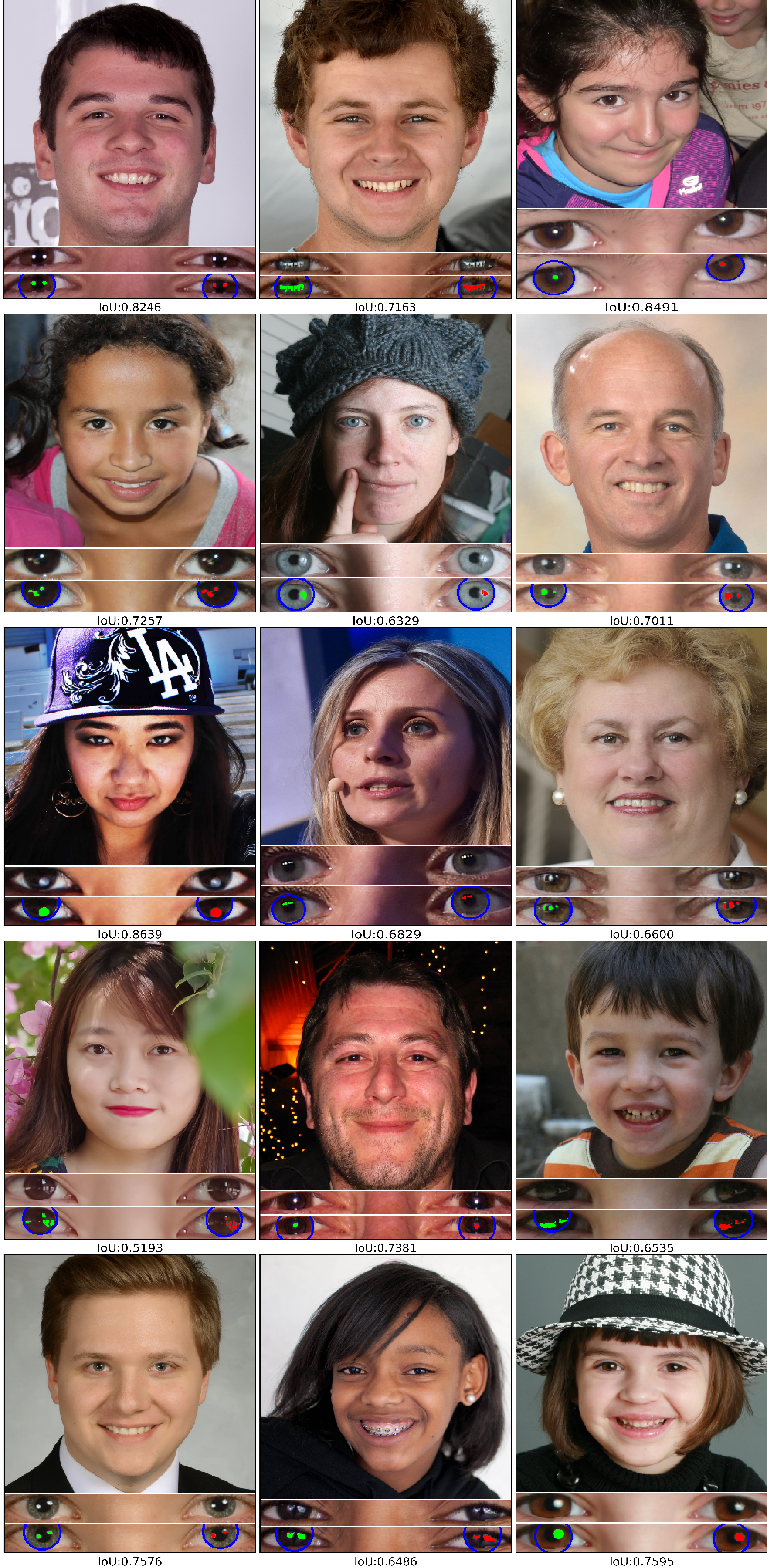}
     \caption{\em More analysis examples of real human eyes.}
     \label{fig:real}
 \end{figure}

 \begin{figure}[t]
     \centering
     \includegraphics[width=.5\textwidth]{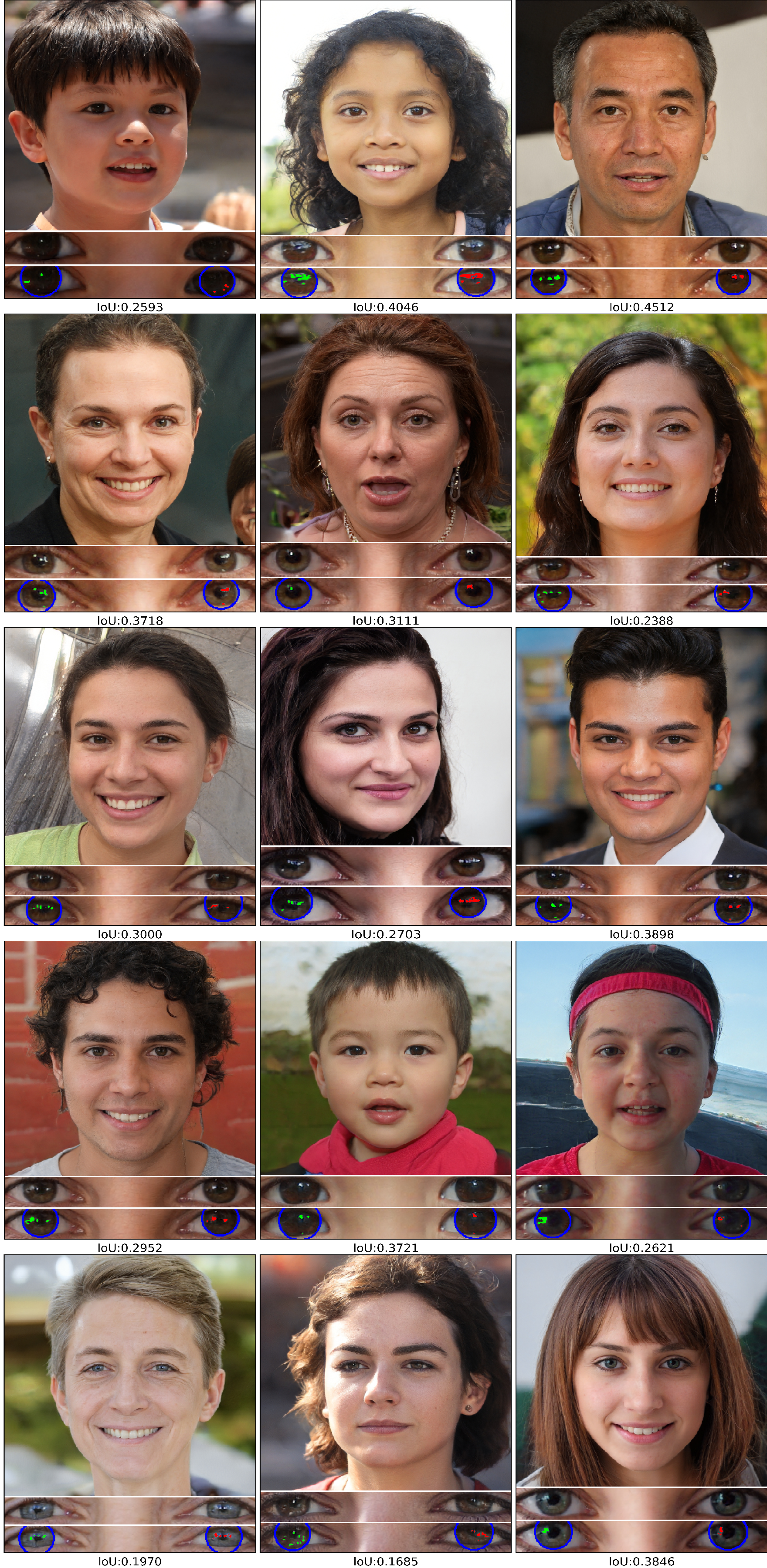}
     \caption{\em More analysis examples of GAN synthesized eyes.}
     \label{fig:fake}
 \end{figure}
\fi
\end{document}